\documentclass[a4paper,11pt]{article}

\usepackage{booktabs}

\usepackage{lineno}
\usepackage[colorlinks=false]{hyperref}
\usepackage{bm}
\usepackage{graphicx}
\usepackage{amssymb}
\usepackage{amsmath}
\usepackage{tabularx}
\usepackage{natbib}
\usepackage{floatrow}
\usepackage{caption}
\usepackage{multirow}
\usepackage{amsmath}
\usepackage{amsfonts}
\usepackage{amssymb}
\usepackage{xcolor}
\usepackage{subfig}
\usepackage{graphicx,amsmath,bm}
\usepackage[binary-units=true]{siunitx}
\usepackage{array}
\usepackage{authblk}
\usepackage{rotating}
\usepackage{enumerate}
\usepackage{ragged2e}

\usepackage[left=15mm,right=15mm,top=1.5cm,bottom=1.5cm,includeheadfoot]{geometry}
\setcitestyle{square,numbers}
\begin{document}
	
	\title{Fast Physics-Driven Untrained Network for Highly Nonlinear Inverse Scattering Problems}
	
	\author[1]{Yutong~Du}
	\author[1]{Zicheng~Liu}
	\author[2]{Yi~Huang}
    \author[3]{Bazargul~Matkerim}
	\author[1]{Yali~Zong}
	\author[1]{Bo~Qi}
	\author[1]{Peixian~Han}
	
	\affil[1]{\scriptsize Department of Electronic Engineering, Northwestern Polytechnical University, Xi'an 710029, China}
	\affil[2]{\scriptsize Department of Physics and Technology, UiT The Arctic University of Norway, 9037, Tromsø, Norway}
	\affil[3]{\scriptsize Department of Computer Science, Al-Farabi Kazakh National University, Almaty 050040, Kazakhstan}
	\maketitle
	
	\abstract{
		Untrained neural networks (UNNs) offer high-fidelity electromagnetic inverse scattering reconstruction but are computationally limited by high-dimensional spatial-domain optimization. We propose a Real-Time Physics-Driven Fourier-Spectral (PDF) solver that achieves sub-second reconstruction through spectral-domain dimensionality reduction. By expanding induced currents using a truncated Fourier basis, the optimization is confined to a compact low-frequency parameter space supported by scattering measurements. The solver integrates a contraction integral equation (CIE) to mitigate high-contrast nonlinearity and a contrast-compensated operator (CCO) to correct spectral-induced attenuation. Furthermore, a bridge-suppressing loss is formulated to enhance boundary sharpness between adjacent scatterers. Numerical and experimental results demonstrate a 100-fold speedup over state-of-the-art UNNs with robust performance under noise and antenna uncertainties, enabling real-time microwave imaging applications.}
	
	\section{Introduction}
	%
	%
	%
	%
	Electromagnetic inverse scattering problems (ISPs) \cite{chen2018computationalEMIS} , which aim to reconstruct the constitutive parameters (\emph{e.g.}, permittivity and conductivity) of unknown objects from measured scattered fields, have remained a cornerstone of microwave research for decades. These techniques are vital for a broad range of applications, including biomedical imaging, non-destructive evaluation, geophysical exploration, and target identification \cite{Qin2022medical,Yan2025medical,An2024CompoMat,liu2025computational,Qin2024Optical}. However, the solution of ISPs is inherently challenging due to their severe nonlinearity and ill-posedness, which are primarily caused by the complex interactions and multiple scattering effects between the incident waves and the scatterers.

    Various deterministic and stochastic methods have been developed to address these challenges. Classical deterministic methods, such as the  contrast source inversion (CSI)\cite{peter1997CSI,richard2001CSI}, the Born iterative method (BIM)\cite{wang1989BIM}, and the distorted Born iterative method (DBIM)\cite{chew1990DBIM}, typically rely on iterative optimization to minimize the discrepancy between measured and simulated data. While effective, these methods are often susceptible to local minima and require carefully chosen initial guesses.
    Subspace-based methods, such as the subspace-based optimization method (SOM) \cite{chen2010SOM}, have been introduced to mitigate nonlinearity by dividing the induced current into deterministic and redundant components. Despite their robustness, these traditional methods often suffer from high computational costs, particularly for large-scale or high-contrast scenarios.
    
    The emergence of deep learning (DL) has ushered in a new paradigm for solving ISPs. Supervised learning approaches \cite{wei2018BPS,Li2018DeepNIS,wei2019PhaNN,Liu2022PhaGuideNN,Liu2022SOMnet,Zhang2023CSINet,Du2025QuaDNN} leverage large datasets to train neural networks that can map scattered fields directly to object profiles. Once trained, these models offer near-instantaneous reconstruction. However, their performance is heavily dependent on the quality and diversity of the training data, and they often struggle with generalization when faced with experimental data or scenarios outside the training distribution.
    
    To circumvent the need for massive datasets, Untrained neural networks (UNNs) have gained significant attention. The solvers including the recently proposed uSOM-Net \cite{Song2022uSOM} and physics-driven neural networks (PDNNs) \cite{Du2025PDNN} treat the network weights as optimization variables and use the physics of the scattering process (expressed via integral equations) as the loss function. While UNNs avoid generalization issues and provide high-fidelity results, their primary drawback is computational efficiency. Existing untrained solvers typically operate in the spatial domain, requiring thousands of iterations over high-dimensional pixel grids, which often results in reconstruction times ranging from tens to hundreds of seconds. Such delays limit their utility in real-time or high-throughput microwave imaging applications.
    
    In this paper, we propose a physics-driven Fourier-spectral (PDF) neural network solver designed to achieve sub-second reconstruction without sacrificing imaging accuracy. The core innovation lies in the synergy between spectral-domain dimensionality reduction and physics-informed deep learning. By expanding the induced currents using a truncated Fourier basis, we exploit the fact that scattering measurements primarily capture low-frequency information. This allows the optimization to occur in a significantly reduced parameter space. An UNN is then utilized to parameterize the updates of these Fourier coefficients, enabling efficient GPU-accelerated optimization.
    
    The main contributions of this work are summarized as follows:
    
    (1) Fourier-Spectral Expansion: We introduce a truncated Fourier basis to represent induced currents. This approach effectively filters out high-frequency noise and reduces the dimensionality of the optimization problem, leading to a substantial speedup in convergence.
    
    (2) Integration of CIE Physics: To address high nonlinearity, the contraction integral equation (CIE) \cite{Zhong2016NIE,Zhong2019CIE,Xu2020FBECIE} is incorporated into the PDF solver. The CIE exhibits better convergence properties than the standard Lippmann-Schwinger equation, particularly for high-contrast objects.
    
    (3) Contrast-compensated operator (CCO): A novel post-processing operator is designed to correct the non-uniformity and boundary attenuation typical of spectral-domain reconstructions, significantly improving the quantitative accuracy of the permittivity profile.
    
    (4) Bridge-Suppressing Loss: We propose a specialized loss function to enhance the separation and boundary sharpness between closely spaced or partially overlapping scatterers, addressing a common resolution bottleneck in ISPs.
    
    In this paper, scalar variables are denoted in italic font and vector (or matrix) variables in bold. Superscript ``H" stands for conjugate operation. The Euclidean length is denoted by $\Vert\cdot\Vert$, and the diagonalization operator is represented as diag($\cdot$).
    
    \section{Physics-Based Formulation of the Inverse Scattering Problem}
    \label{sec:formulateISPs}
\begin{figure}
    	\centering
    	\includegraphics[width = 0.4\linewidth]{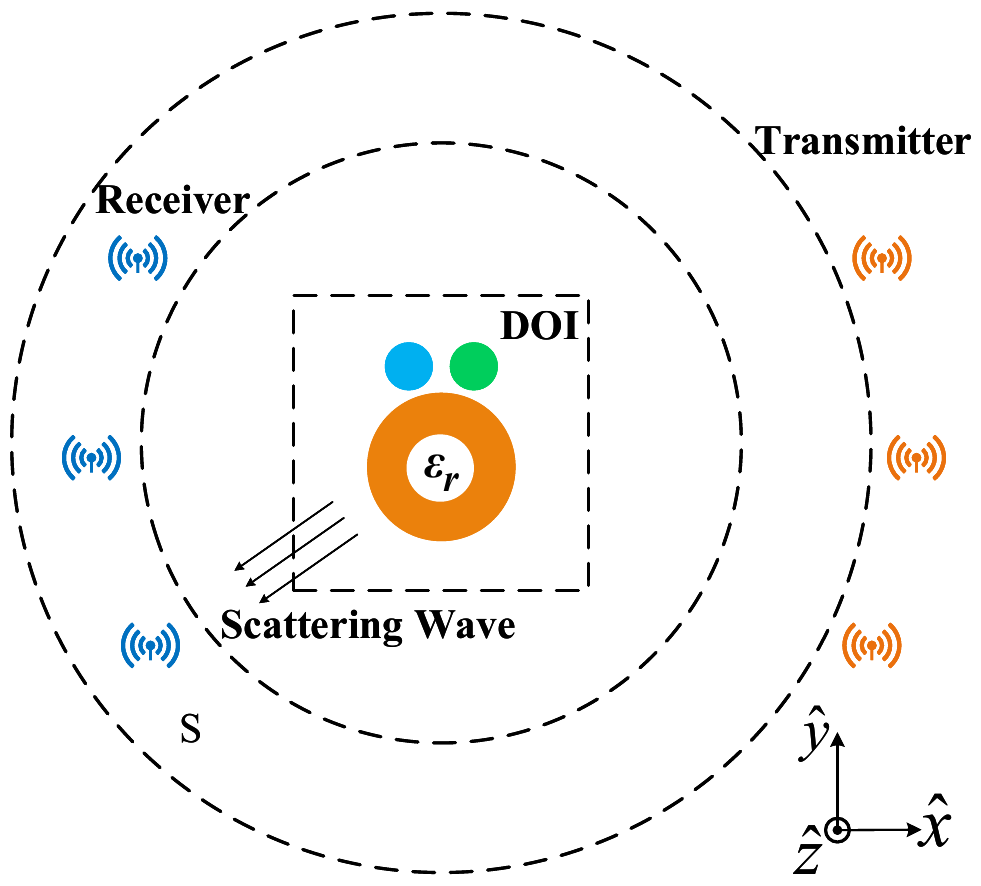}
    	\caption{Schematic of the 2-D inverse scattering configuration.}
    	\label{fig:ISPs}
    \end{figure}

    The schematic diagram of the two-dimensional ISPs is illustrated in Figure~\ref{fig:ISPs}, where the domain of interest (DOI) is sequentially illuminated by transverse magnetic (TM) waves from the ${N_{i}}$ transmitters. The DOI is discretized into an $M_1 \times M_2$ uniform grid for numerical calculation. Scattered fields are collected by ${N_{s}}$ receivers in the space S to reconstruct the electrical properties (e.g., relative permittivity, conductivity, etc.) of the objects within the DOI. 
    \subsection{Lippmann-Schwinger Integral Model}
    \label{subsec:LSIE}
    The electromagnetic scattering process is characterized by the state and data equations. The state equation describes the coupling between the incident field and the scatterers, as well as the multiple scattering effects within the objects,
    \begin{equation}
    	\mathbf{E}^\text{tot}(\mathbf{r}) = \mathbf{E}^\text{inc}(\mathbf{r}) + k_0^2\int_\text{DOI}g(\mathbf{r},\mathbf{r}^\prime)\mathbf{J}(\mathbf{r}^\prime)d\mathbf{r}^\prime, \,\mathbf{r}\in\text{DOI},
    	\label{eq:stateEqu}
    \end{equation}
    where $\mathbf{E}^\text{tot}(\mathbf{r})$ and $\mathbf{E}^\text{inc}(\mathbf{r})$ are the total and incident electric fields at position ${\mathbf{r}}$ within the DOI. $k_0$ denotes the free-space wavenumber. $g$ is the scalar Green's function. The induced current source $\mathbf{J}$ is defined as $\mathbf{J}(\mathbf{r}^\prime)=\boldsymbol{\chi}(\mathbf{r}^\prime)\mathbf{E}^\text{tot}(\mathbf{r}^\prime)$, where $\boldsymbol{\chi}(\mathbf{r}^\prime) = \epsilon_r(\mathbf{r}^\prime)-1$ represents the contrast and $\epsilon_r(\mathbf{r}^\prime)$ is the relative permittivity. 
    
    The data equation quantifies the scattered fields collected by the receivers, 
    \begin{equation}
    	\mathbf{E}^\text{sca}(\mathbf{r}) = k_0^2\int_\text{DOI}g(\mathbf{r},\mathbf{r}^\prime)\mathbf{J}(\mathbf{r}^\prime)d\mathbf{r}^\prime, \,\mathbf{r}\in\text{S}.
    	\label{eq:dataEqu}
    \end{equation} 
    Denoting $\mathbf{G}_\text{D}$ and $\mathbf{G}_\text{S}$ as the integral operators for the state and data equations, respectively, \eqref{eq:stateEqu} and \eqref{eq:dataEqu} can be expressed in operator form,
    \begin{equation}
    	\mathbf{E}^\text{tot} = \mathbf{E}^\text{inc} + \mathbf{G}_\text{D}\boldsymbol{\chi}\mathbf{E}^\text{tot}, 
    	\label{eq:GreenOperatorStateEqu}
    \end{equation}
    \begin{equation}
    	\mathbf{E}^\text{sca} = \mathbf{G}_\text{S}\boldsymbol{\chi}\mathbf{E}^\text{tot}.
    	\label{eq:GreenOperatorDataEqu}
    \end{equation}
    Combining \eqref{eq:GreenOperatorStateEqu} and \eqref{eq:GreenOperatorDataEqu} yields the nonlinear mapping from the contrast $\boldsymbol{\chi}$ to the scattered field $\mathbf{E}^\text{sca}$, 
    \begin{equation}
    	\mathbf{E}^\text{sca} = \mathbf{G}_\text{S}(\mathbf{I}-\boldsymbol\chi\mathbf{G}_\text{D})^{-1}\boldsymbol\chi\mathbf{E}^\text{inc}.
    	\label{eq:mapping}
    \end{equation}
    The term $(\mathbf{I}-\boldsymbol\chi\mathbf{G}_\text{D})^{-1}$ can be expanded as a Neumann series $ \sum_{n=0}^\infty(\boldsymbol\chi\mathbf{G}_\text{D})^n$, which provides physical insight into the multiple-scattering mechanism. For high-contrast scatterers, the magnitude of $\boldsymbol\chi\mathbf{G}_\text{D}$ increases, causing the series to converge slowly or diverge, thereby significantly exacerbating the nonlinearity and ill-posedness of the ISPs.
    
    \subsection{Contraction Integral Equation Model}
    \label{subsec:CIE}
    To alleviate the nonlinearity of ISPs, the contraction integral equation (CIE) model \cite{Zhong2019CIE} reformulates the state equation by introducing an auxiliary variable $\boldsymbol{R}$ to rescale the contrast-induced interaction,
    \begin{equation}
    	\beta\mathbf{J}=\boldsymbol{R}(\beta\mathbf{J}+\mathbf{E}^\text{inc}+\mathbf{G}_\text{D}\mathbf{J})
    	\label{eq:CIEStateEqu}
    \end{equation}
    where the modified contrast 
    \begin{equation}
    	\boldsymbol{R}=\frac{\beta\boldsymbol\chi}{\beta\boldsymbol\chi+1}
    	\label{eq:R}
    \end{equation}
    $\beta$ is a hyperparameter. For physically admissible media (where the real and the imaginary part of $\boldsymbol\chi$ are nonnegative), choosing $\beta$ with a positive real part ensures that the mapping $\boldsymbol\chi\mapsto\boldsymbol R$ is contractive. Consequently, the data equation is rewritten in terms of the modified contrast
    \begin{equation}
    	\mathbf{E}^\text{sca} = \frac{1}{\beta}\mathbf{G}_\text{S}[\mathbf{I}-\boldsymbol{R}(\mathbf{I}+\frac{1}{\beta}\mathbf{G}_\text{D})]^{-1}\boldsymbol{R}\mathbf{E}^\text{inc}
    	\label{eq:CIEDataEqu}
    \end{equation}
    By bounding the norm of the interaction operator $\boldsymbol{R}(\mathbf{I}+\frac{1}{\beta}\mathbf{G}_\text{D})$, the CIE transforms the strongly nonlinear mapping into a weakly nonlinear one, enhancing reconstruction stability for high-contrast targets.
    
    \subsection{Fourier Bases-Expansion}
    \label{subsec:FBE}
    
    While the CIE alleviates nonlinearity, the unknown contrast remains a high-dimensional spatial function. Directly optimizing over a dense spatial grid is computationally demanding and often redundant, as the scattering information captured in the measured data is inherently band-limited due to the low-pass nature of the Green’s function. To exploit this spectral compressibility, we represent the induced current $\mathbf{J}$ using a truncated discrete Fourier basis expansion (FBE)\cite{Xu2020FBECIE}
    \begin{equation}
        \mathbf{J} = [\mathbf{F}_1, \mathbf{F}_2, \dots, \mathbf{F}_{M_0}] \boldsymbol{\alpha},
    \end{equation}
    where $\mathbf{F}_k$ are vectorized spatial-domain Fourier basis functions and $\boldsymbol{\alpha}\in\mathbb{C}^{M_0}$ is the vector of corresponding expansion coefficients. 
    
    In this implementation, we retain only the low-frequency components by selecting four $M_F \times M_F$ blocks located at the corners of the 2-D discrete Fourier transform (DFT) spectrum (corresponding to the lowest spatial frequencies), resulting in $M_0 = 4M_F^2$ retained modes. This spectral parameterization substantially reduces the dimensionality of the unknown space, thereby decreasing computational costs and providing an inherent regularization effect that filters out high-frequency noise. For notational convenience, the 2-D DFT and inverse DFT (IDFT) operations are hereafter denoted as $F_T(\cdot)$ and $F^{*}_T(\cdot)$, respectively,  such that $\boldsymbol\alpha = F_T(\mathbf{J})$ and $\mathbf{J}=F^{*}_T(\mathbf{\boldsymbol\alpha})$.
    
    \subsection{Contrast-Compensated Operator}
    
    \label{subsec:CCOperator}
    \begin{figure}
    	\centering
    	\includegraphics[width = .76\linewidth]{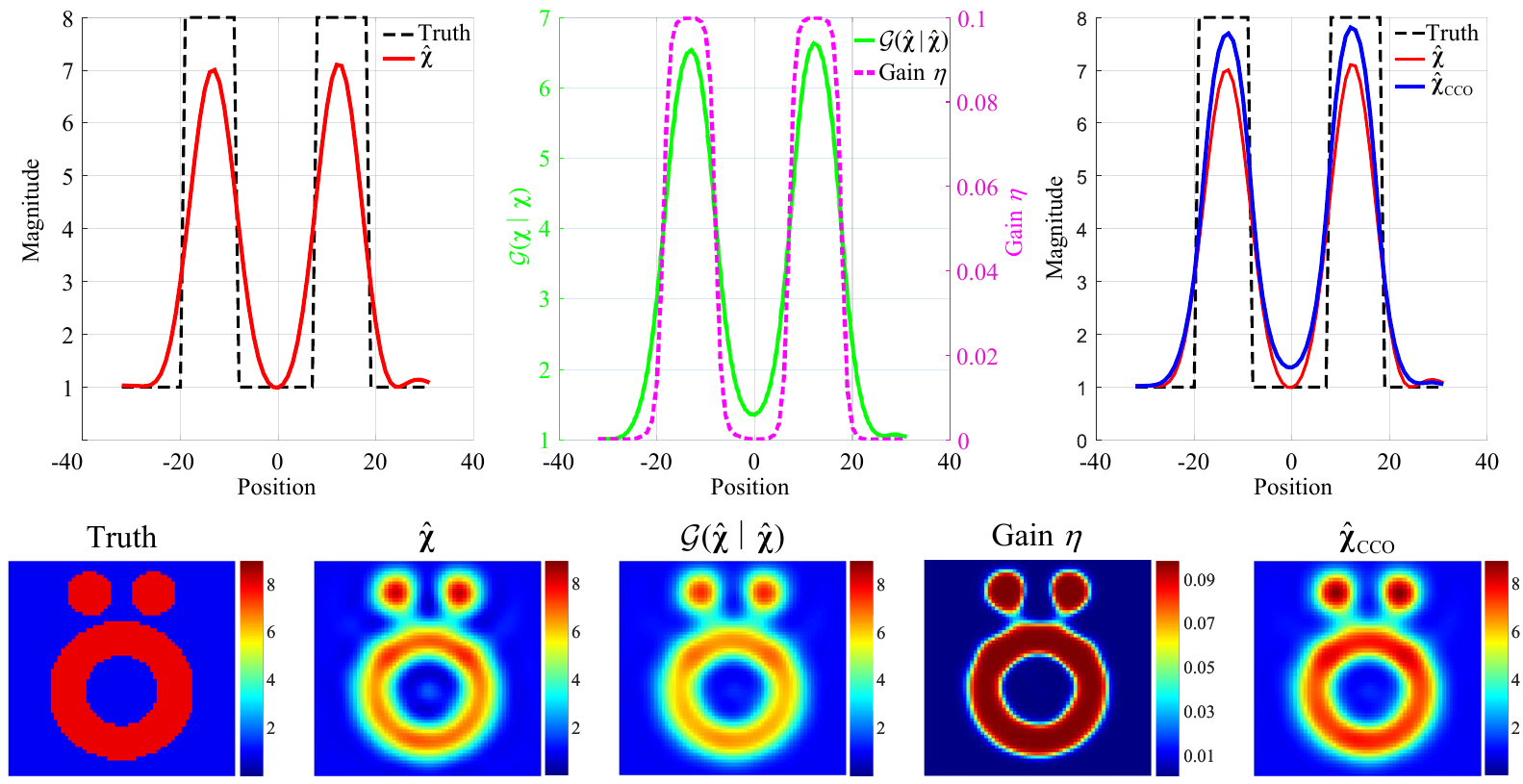}
    	\caption{Conceptual workflow of the contrast-compensated operator (CCO) demonstrating edge-restoration through self-guided projection.}
    	\label{fig:example}
    \end{figure}
    
    As illustrated in Figure~\ref{fig:example}, the contrast profile reconstructed by the network typically exhibits a ``edge roll-off" effect: the reconstructed values are highest in the object's interior but gradually diminish toward its boundaries. This systematic attenuation is a fundamental consequence of the truncated Fourier basis representation introduced in Section~\ref{subsec:FBE}. By retaining only low-frequency components, the reconstruction essentially undergoes a low-pass filtering process that smooths sharp dielectric interfaces and reduces the peak amplitude of the contrast, particularly for high-contrast objects.
    
    To compensate for this spectral-domain-induced attenuation, a contrast-compensated operator (CCO) is applied as a post-processing step. The CCO first performs a self-guided projection \cite{He2010GuideFilter, He2013GuideFilter} to refine the initial estimate $\hat{\chi}$, 
    \begin{equation}
    \hat{\boldsymbol\chi}_\text{CCO}=(1+\eta)\,\mathcal{G}(\hat{\boldsymbol\chi}\mid\hat{\boldsymbol\chi})
    +\eta\,\hat{\boldsymbol\chi},
    \label{eq:CCO}
    \end{equation}
    where $\mathcal{G}(\cdot|\cdot)$ denotes the self-guided projection operator. Unlike conventional convex averaging, this operator utilizes a gain-modulated projection where the compensation is adaptively controlled by a contrast-aware gain factor $\eta$,
    \begin{equation}
    \eta(\hat{\boldsymbol\chi})
    =\eta_{\max}\,
    \sigma\!\left(
    \frac{|\hat{\boldsymbol\chi}|-\tau}{\delta}
    \right).
    \label{eq:contrastGain}
    \end{equation}
    Here, $\sigma(\cdot)$ denotes the logistic sigmoid function. The threshold $\tau$ controls the activation of contrast compensation, $\eta_{\max}$ limits the maximum compensation strength, and $\delta$ controls the slope of the sigmoid function, thereby determining the smoothness of the transition near the threshold. 
    
    In this work, we set $\tau=3$, $\eta_{\max}=0.1$, and $\delta=0.5$. Consequently, the gain $\eta$ remains negligible in low-contrast regions where spectral truncation has a minimal impact, but increases smoothly in regions of high reconstructed contrast where systematic attenuation is most prevalent. This mechanism provides a spatially selective compensation that restores the peak permittivity values without introducing global amplification or artifacts in the background. The effectiveness of the CCO is visually demonstrated in Figure~\ref{fig:example} and quantitatively validated in Section~\ref{subsec:ContrCCOandBri}.

    \section{Physics-Driven Fourier-Spectral Neural Network Solver}
    \label{sec:PDFsolver}
    Under the CIE formulation, the induced current (and the corresponding contrast source) exhibits reduced spatial oscillations and is predominantly characterized by its low-frequency spectral components. Motivated by this spectral compressibility, we represent the unknowns using a truncated Fourier basis expansion, retaining only a subset of low-frequency coefficients. Consequently, the reconstruction task is reformulated from the high-dimensional recovery of spatial variables to the estimation of a compact coefficient vector $\boldsymbol{\alpha}$, which is optimized through a physics-driven neural network.
    
    \begin{figure*}
    	\centering
    	\includegraphics[width = \linewidth]{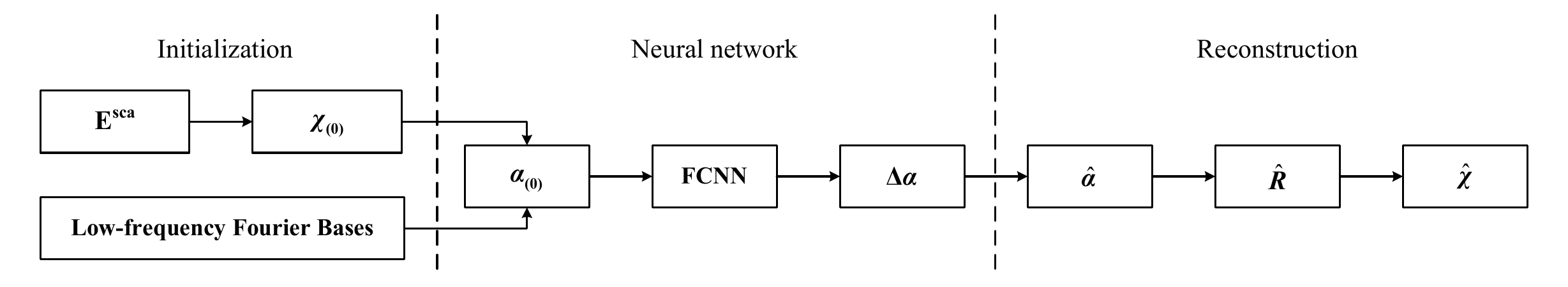}
    	\caption{Computational pipeline of the proposed PDF solver, divided into initialization, neural network optimization, and profile reconstruction.}
    	\label{fig:overview}
    \end{figure*}
    
    The proposed PDF solver consists of three-parts as shown in Figure~\ref{fig:overview}. The framework utilizes the back propagation (BP) method \cite{tsili1998BP} to acquire an initial estimation ${\boldsymbol\chi}_{(0)}$ from the measured scattered fields, and constructs a truncated low-frequency Fourier basis. Then, the initial Fourier coefficients $\boldsymbol{\alpha}_{(0)}$ are derived and fed into a deep neural network to learn a corrective update $\Delta \boldsymbol{\alpha}$. At the end of iteration, the obtained coefficients $\boldsymbol{\alpha}$ are used to determine the operator $\boldsymbol{R}$, ultimately yielding an accurate reconstruction of the contrast ${\boldsymbol\chi}$.
    
    \subsection{Network architecture}
    \label{subsec:nnArch}
    
    \begin{figure}
    	\centering
    	\includegraphics[width = .6\linewidth]{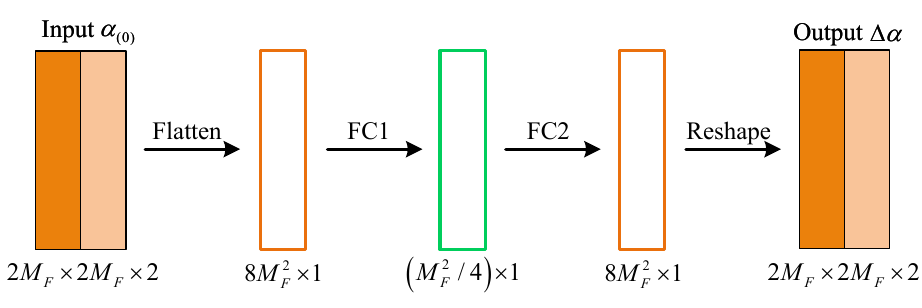}
    	\caption{The network architecture of the proposed solver, where the input consists of two channels representing the real and imaginary parts of the initial Fourier coefficients by the 2-D discrete Fourier transform.}
    	\label{fig:nnArch}
    \end{figure}
    
    The architecture of the proposed Physics-Driven Fourier-spectral (PDF) solver is illustrated in Figure~\ref{fig:nnArch}. The solver is implemented as a fully connected neural network (FCNN) designed to predict a corrective update $\Delta \boldsymbol\alpha$ for the Fourier coefficients, starting from an initial estimate $\boldsymbol\alpha_{(0)}$. The updated coefficients are then given by $\hat{\boldsymbol\alpha} = \boldsymbol\alpha_{(0)} + \Delta \boldsymbol\alpha$.
    
    The network input consists of two channels representing the real and imaginary parts of the initial Fourier coefficients. Specifically, an initial contrast estimate ${\boldsymbol\chi}_{(0)}$ is first computed using a fast back-propagation (BP) imaging algorithm, and, based on ${\boldsymbol\chi}_{(0)}$, the modified contrast $\boldsymbol R_{(0)}$ is then computed according to \eqref{eq:R}. Finally, the initial Fourier coefficients $\boldsymbol{\alpha}_{(0)}$ are set as the CSI solution where the objective function is 
    \begin{equation}
    L^\text{CSI}(\hat{\boldsymbol\alpha})=\frac{||\mathbf{G}_\text{S}F^{*}_T(\hat{\boldsymbol\alpha})-\mathbf{E}^\text{sca}||_{2}^{2}}{||\mathbf{E}^\text{sca}||_{2}^{2}}+\frac{||\hat{\boldsymbol{R}}\mathbf{E}^\text{inc}+[\hat{\boldsymbol{R}}\mathbf{G}_\text{D}-\beta(\mathbf{I}-\hat{\boldsymbol{R}})] F^{*}_T(\hat{\boldsymbol\alpha})||_{2}^{2}}{||\mathbf{E}^\text{inc}||_{2}^{2}}.
    \end{equation}
    To avoid the low optimization efficiency of CSI, the single-step gradient update starting from zero is chosen as $\boldsymbol{\alpha}_{(0)}$ and have been shown effective from the analysis in Section \ref{sec:numAna}.

    \subsection{Loss Function}
    \label{subsec:lossFunc}
    The PDF solver utilizes a composite loss function to ensure physical consistency and spatial regularization,
    \begin{equation}
        L=L^\mathrm{Physics} + \lambda_1 L^\mathrm{Bound} + \lambda_2 L^\mathrm{TV} + \lambda_3 L^\mathrm{Bridge}
        \label{eq:lossfun}
    \end{equation} 
    where $\lambda_1, \lambda_2, \text{ and } \lambda_3$ are weight hyperparameters. The physics-consistency loss $L^\mathrm{Physics}$ is a summation of the state-consistency loss $L^\mathrm{State}$ and the data-misfit loss $L^\mathrm{Data}$,
    \begin{equation}
    	L^\mathrm{Physics}=L^\mathrm{State}+L^\mathrm{Data},
    	\label{eq:LPhysics}
    \end{equation}
    where 
    \begin{equation}
    	L^\mathrm{State}=\frac{||\hat{\boldsymbol{R}}\mathbf{E}^\text{inc}+[\hat{\boldsymbol{R}}\mathbf{G}_\text{D}-\beta(\mathbf{I}-\hat{\boldsymbol{R}})] F^{*}_T(\hat{\boldsymbol\alpha})||_{2}^{2}}{||\mathbf{E}^\text{inc}||_{2}^{2}},
    	\label{eq:lossState}
    \end{equation}
    ensures that the reconstructed current satisfies the CIE-modified state equation, and
    \begin{equation}
    	L^\mathrm{Data}=\frac{||\mathbf{G}_\text{S}F^{*}_T(\hat{\boldsymbol\alpha})-\mathbf{E}^\text{sca}||_{2}^{2}}{||\mathbf{E}^\text{sca}||_{2}^{2}}.
    	\label{eq:lossData}
    \end{equation}
    minimizes the residual between the measured and simulated scattered fields. $L^\mathrm{Bound}$ is the physical boundary constraint defined as
    \begin{equation}
      L^\mathrm{Bound}=||\text{ReLU}(-\mathbf{Re\{\hat{\boldsymbol{\chi}}}\})||_{2}^{2},
      \label{eq:LBound}
    \end{equation}
    which enforces the non-negativity of the real part of the contrast, which is physically required for common dielectric media. $\hat{\boldsymbol{R}}$ and $\hat{\boldsymbol{\chi}}$ are the solutions from CIE.
    
    $L^{\text{TV}}$ is the total-variation (TV) regularization term, promoting piecewise smoothness and suppresses noise-induced artifacts while preserving sharp object boundaries, defined as
    \begin{equation}
    	\label{eq:LTV}
    	L^{\text{TV}}=\sum_{i,j}\sqrt{(\hat{\boldsymbol{\chi}}_{i,j+1}-\hat{\boldsymbol{\chi}}_{i,j})^2+(\hat{\boldsymbol{\chi}}_{i+1,j}-\hat{\boldsymbol{\chi}}_{i,j})^2},
    \end{equation}
    
    $L^{\text{Bridge}}$ is the term designed to penalize artificial ``bridges" between closely spaced scatterers,
    \begin{equation}
    	\label{eq:LBridge}
    	L^{\text{Bridge}}=\sum_{i,j}\sigma(\frac{|\hat{\boldsymbol{\chi}}_{i,j}|-\tau_B}{\tau_B}) \cdot \exp\left(-\|\nabla|\hat{\boldsymbol{\chi}}_{i,j}|\|_{2}^{2}\right).
    \end{equation}
    where $\sigma(\cdot)$ is the sigmoid function and $\tau_B$ is a threshold parameter empirically set as 0.5. This loss penalizes regions with high amplitude but low gradients, effectively mitigating incorrect connections between neighboring objects. 
    
    In this paper, the hyperparameter $\lambda_1$, $\lambda_2$ and $\lambda_3$ are empirically set as $1\times10^{-3}$, $1\times10^{-5}$ and $1\times10^{-5}$, which are shown reasonable from the analysis in Section \ref{sec:numAna}.
    
    \subsection{Training settings}
    \label{subsec:trainSet}
    The PDF solver is implemented in a deep learning framework and optimized using the Adam optimizer with a learning rate of $1 \times 10^{-2}$. All experiments were performed on a workstation equipped with an NVIDIA GeForce RTX 4090 GPU and 128 GB of RAM. By leveraging the spectral-domain reduction and GPU acceleration, the solver achieves high-fidelity reconstructions with sub-second inference times.
    
    \section{Numerical Analysis}
    \label{sec:numAna}
    To evaluate the performance of the proposed physics-driven Fourier-spectral (PDF) solver, extensive simulations and benchmark tests are conducted. The domain of interest (DOI) is defined as a $1.5\, \text{m} \times 1.5\, \text{m}$ square region, uniformly discretized into a $64 \times 64$ grid. Scattered fields are generated using the method of moments (MoM) \cite{Gibson2021MoM} at a frequency of 400 MHz. The measurement configuration consists of 36 transmitters and 36 receivers equally spaced on a circle of radius $20\lambda$, where $\lambda$ denotes the free-space wavelength.
    
    \subsection{Multi-Parameter Sensitivity Analysis: $\beta$, $M_F$, and $k$}
    \label{subsec:influIter}
    
    \begin{figure}[!t]
    	\centering
    	\includegraphics[width = .52\linewidth]{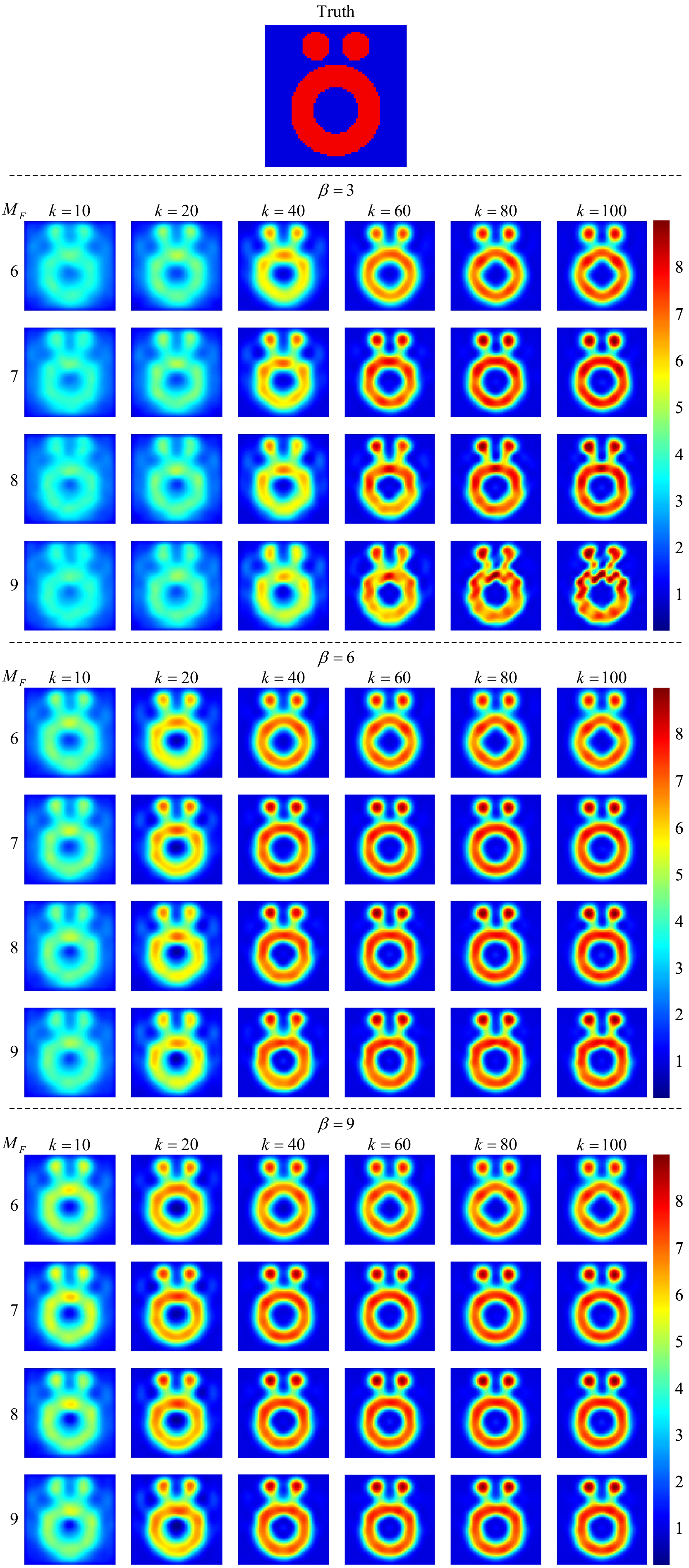}
    	\caption{A comprehensive parameter analysis by fixing $\beta$ and varying the iteration number $k$ and $M_F$.}
    	\label{fig:iterDiffLFandBeta}
    \end{figure}
    \begin{figure}[!t]
    	\centering
    	\includegraphics[width = .49\linewidth]{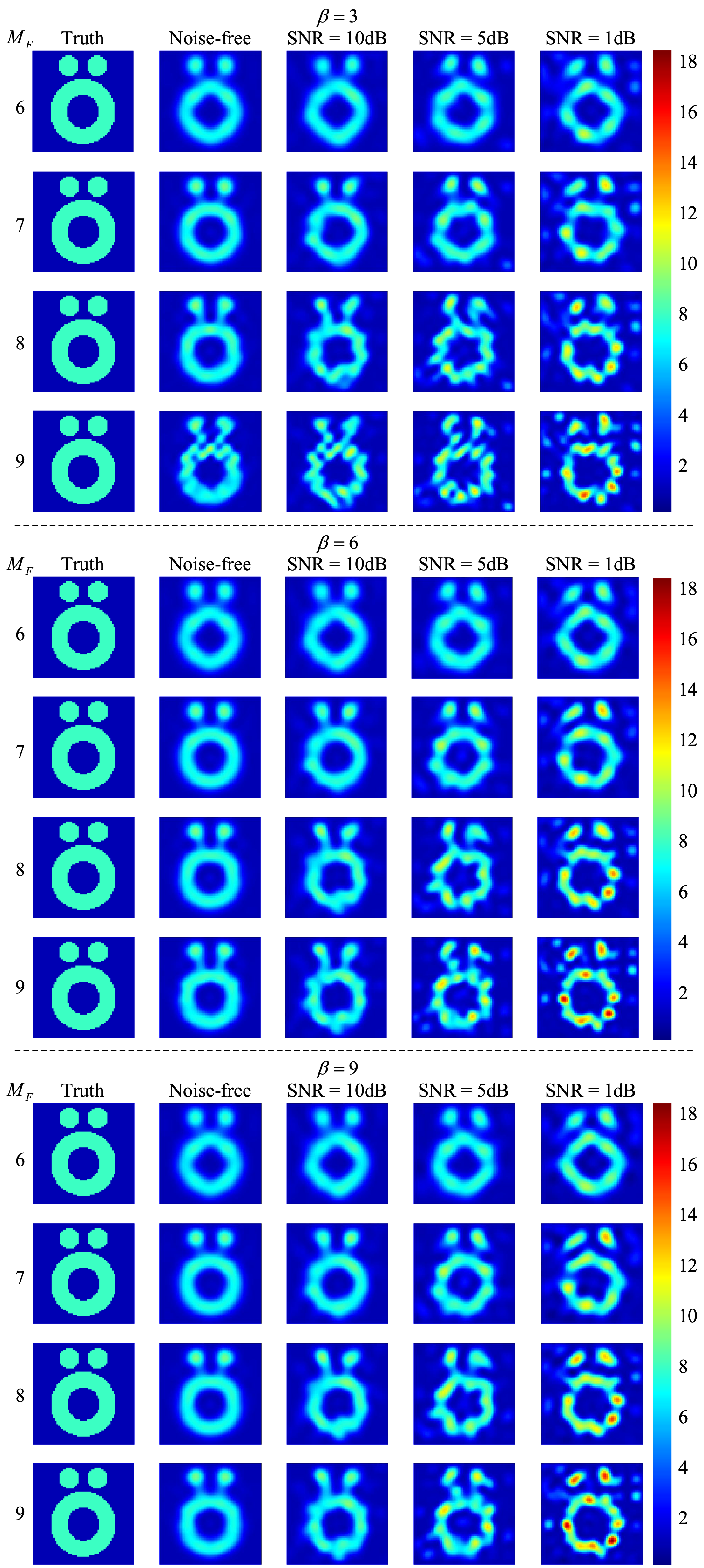}
    	\caption{The reconstruction results of different hyperparameter $\beta$ and $M_F$ under the condition of different noise levels.}
    	\label{fig:noiseDiffLFandBeta}
    \end{figure}
    The performance of the PDF solver is governed by the interplay between three key parameters: the iteration number $k$, the contraction hyperparameter $\beta$, and the Fourier truncation order $M_F$, which represents the optimization depth, the degree of nonlinearity mitigation, and the spectral representation capacity, respectively. These parameters are intrinsically coupled and their optimal values should be determined through systematic sensitivity analysis.
    
    First, study the effects from $k$. As illustrated in Figure~\ref{fig:iterDiffLFandBeta}, the reconstruction quality for the Austria profile exhibits monotonic improvement as the iteration number $k$ increases from 10 to 100. This trend is consistent across various spectral orders and contraction strengths, demonstrating the inherent stability of the physics-driven optimization framework. Notably, the solver does not exhibit divergent behavior even at high iteration counts, suggesting that the Fourier-spectral parameterization provides a natural regularization that prevents the optimizer from fitting high-frequency noise. 
    
    The hyperparameter $\beta$ is critical for accelerating convergence in the presence of strong multiple scattering. A clear performance gain is observed when increasing $\beta$ from 3 to 6, where the solver transitions from a slow, artifact-prone reconstruction to a stable, high-fidelity result. This improvement stems from the CIE's ability to transform a strongly nonlinear mapping into a weakly nonlinear one by bounding the interaction operator. However, a saturation effect is observed as $\beta$ is further increased to 9. Beyond a certain threshold, the interaction operator is already sufficiently contractive; further increasing $\beta$ offers marginal utility and may even over-suppress the physical scattering contributions necessary for accurate quantitative reconstruction.
    
    The truncation order $M_F$ dictates the dimensionality of the unknown space. Our results reveal a classic trade-off between representational capacity and ill-posedness. When $M_F=7$, the solver achieves the best balance, successfully preserving the delicate annular structure of the Austria profile with high spatial resolution and contrast uniformity. A lower value of $M_F$ (\emph{e.g.}, $M_F=6$) results in over-smoothed images that fail to capture sharp dielectric interfaces, as the low-pass nature of the basis acts as a spatial filter. Conversely, increasing $M_F$ to 9 leads to significant geometric distortion and a loss of resolution. This occurs because the forward operator in ISPs is inherently low-pass due to the Green’s function; consequently, high-frequency Fourier modes are poorly supported by the far-field measurements. The expanded solution subspace causes remaining multiple-scattering nonlinearity, which are not fully mitigated by the CIE, to be amplified as artifacts. Moreover, the geometric deformation observed at $M_F=9$ is much more severe for $\beta=3$ than for $\beta=6$. This indicates that a stronger contraction parameter effectively acts as a secondary regularizer, suppressing the non-physical amplification of high-frequency modes. Based on this analysis, the configuration of $\beta=6$ and $M_F=7$ is selected as the optimal baseline for the subsequent benchmarks.
    
    \subsection{Robustness to Noise and Comparative Study}
    \label{subsec:influNoise}
    
    The robustness of an inverse scattering solver under adverse signal-to-noise ratio (SNR) conditions is a critical metric for practical microwave imaging. In this subsection, we first analyze how $M_F$ and $\beta$ influence the solver’s noise sensitivity, followed by a comprehensive benchmark against state-of-the-art methods.
    
    \begin{figure}[!t]
    	\centering
    	\includegraphics[width = .49\linewidth]{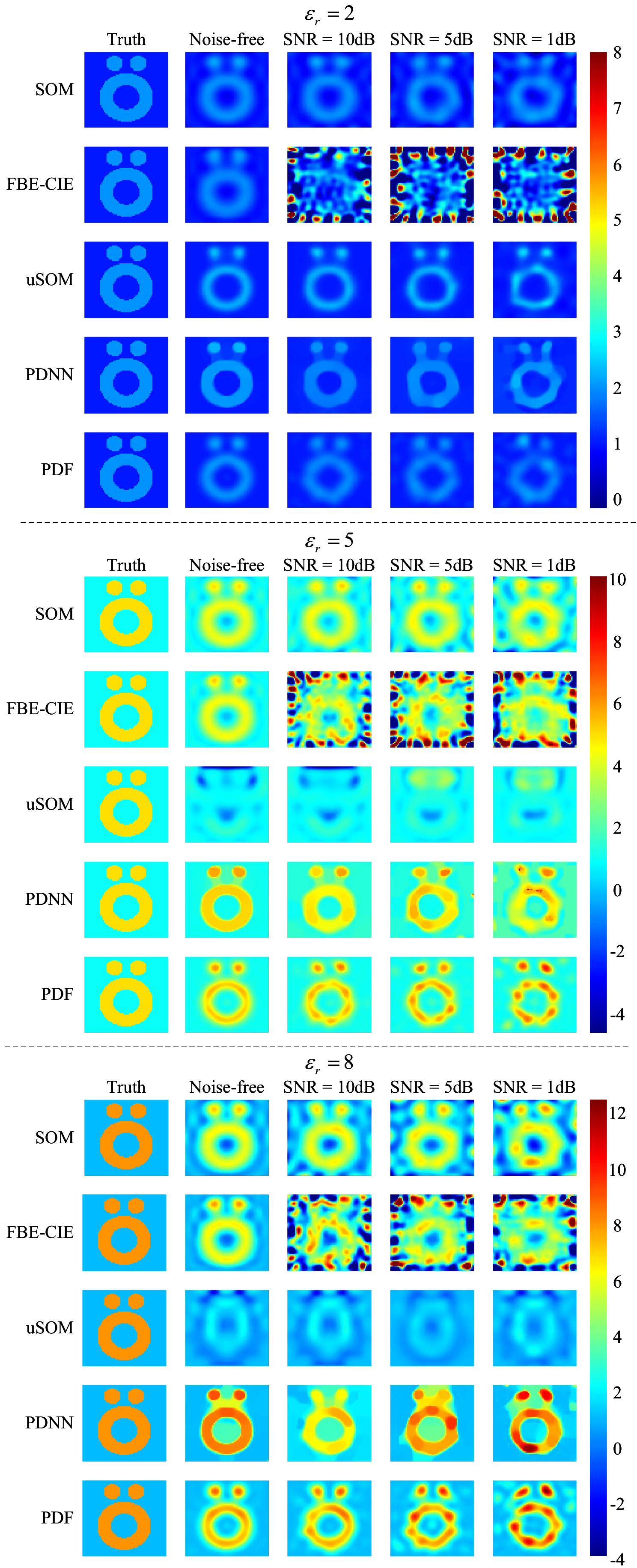}
    	\caption{Reconstructed permittivity profiles for Austria targets under varying contrast levels ($\epsilon_r = 2, 5, 8$) and noise conditions (SNR$ = 10, 5, 1$ dB).}
    	\label{fig:noiseDiffSolvers}
    \end{figure}
    
    \subsubsection{Interplay between noise and model parameters}
    To assess noise robustness, additive white Gaussian noise (AWGN) is injected into the measured scattered fields to simulate real-world measurement uncertainties. As shown in Figure~\ref{fig:noiseDiffLFandBeta}, the noise sensitivity of the PDF solver is closely coupled with its spectral representation. 
    
    For a fixed $\beta$, the solver's sensitivity to noise increases as $M_F$ becomes larger. Smaller $M_F$ values (\emph{e.g.}, $M_F = 6$) yield highly noise-robust reconstructions since the reduced spectral degrees of freedom act as a low-pass filter, effectively suppressing high-frequency noise fluctuations. Conversely, larger $M_F$ values allow the network to fit noise-induced perturbations in the high-frequency components of the spectrum. 
    
    The contraction parameter $\beta$ significantly affects the noise floor of the reconstruction. A moderate value (\emph{e.g.}, $\beta = 6$) provides sufficient contraction to stabilize the iterative updates against noise-driven instabilities. While an overly small $\beta$ leads to severe noise-induced artifacts, an excessively large $\beta$ results in over-smoothing, as it excessively dampens the scattering contributions needed to resolve fine details. Based on this analysis, $\beta = 6$ and $M_F = 7$ are identified as the optimal configuration for balancing resolution and robustness.
    
    \subsubsection{Comparative analysis with baseline solvers}
    The proposed PDF solver is compared with four representative baselines: iterative algorithms (SOM and FBE-CIE) and state-of-the-art untrained neural network solvers (uSOM and PDNN). Tests are conducted across three contrast levels—low ($\epsilon_r = 2$), moderate ($\epsilon_r = 5$), and high ($\epsilon_r = 8$)—under various noise conditions ranging from noise-free to severe SNR$= 1$ dB. The reconstruction results are illustrated in Figure~\ref{fig:noiseDiffSolvers}.
    
    For low-contrast scenarios ($\epsilon_r = 2$), when free of noise, all solvers recover the general geometry of the Austria profile. However, PDNN, SOM, and FBE-CIE frequently introduce non-physical ``bridges" or false connections between the three components of the scatterer. As noise increases, SOM exhibits significant blurring, while FBE-CIE fails entirely at SNR$= 10$ dB due to its sensitivity to high-frequency perturbations. 
    
    For moderate-to-high contrast ($\epsilon_r = 5, 8$) cases, the advantages of the PDF solver become most pronounced in highly nonlinear regimes. The uSOM method fails to provide meaningful reconstructions beyond low contrast. In the high-contrast case ($\epsilon_r = 8$), traditional SOM and FBE-CIE suffer from severe contrast underestimation and background artifacts. While the PDNN maintains sharp edges, it exhibits strong mutual interference between neighboring targets. 
    
    As observed, the PDF solver consistently maintains clear separation between closely spaced targets and preserves the structural integrity of the Austria profile even at SNR$= 1$ dB. Although its boundaries are slightly smoother than PDNN’s due to the Fourier truncation, the PDF solver avoids the large-scale background contamination and geometric distortions that plague other methods under high noise.
    
    \subsubsection{Computational efficiency and inference time}
    A primary bottleneck for UNN-based solvers is the high computational cost of the optimization process. As summarized in Table~\ref{tab:compare}, traditional solvers and existing UNNs (uSOM, PDNN) require between 78 and 321 seconds to complete a single reconstruction. In contrast, the PDF solver consistently achieves high-fidelity results in less than one second ($\sim 0.88$ s). This approximately 100-fold speedup is attributed to the combination of spectral-domain dimensionality reduction. This performance bridge the gap between the high accuracy of iterative solvers and the real-time potential of supervised deep learning, without being constrained by generalization limits.
    
    \begin{table}[!t]
        \centering
        \caption{Comparison of Reconstruction Time (in Seconds) for Different Solvers Across Various Scenarios.}
        \label{tab:compare}
        \resizebox{.6\columnwidth}{!}{
        \begin{tabular}{l|c|c|c|c|c}
            \toprule
            \textbf{} & \multicolumn{1}{c|}{\textbf{Methods}} & \multicolumn{1}{c|}{\textbf{Noise-free}} & \multicolumn{1}{c|}{\textbf{SNR=10dB}} & \multicolumn{1}{c|}{\textbf{SNR=5dB}} & \multicolumn{1}{c}{\textbf{SNR=1dB}} \\  
            \toprule
                            &SOM  &90.13s  &90.42s  &90.10s  &89.82s   \\ 
                            &FBE-CIE  &263.92s  &254.29s  &265.28s  &258.02s   \\
            $\epsilon_r=2$  &uSOM  &78s  &78s  &79s  &78s   \\
                            &PDNN  &251s  &234s  &249s  &250s   \\
                            &PDF   &0.88s  &0.87s  &0.95s  &0.91s   \\
            \midrule
                            &SOM  &90.10s  &89.63s  &90.26s  &90.55s   \\ 
                            &FBE-CIE  &254.22s  &267.16s  &269.08s  &254.89s   \\
            $\epsilon_r=5$  &uSOM  &78s  &78s  &78s  &79s   \\
                            &PDNN  &304s  &276s  &266s  &284s   \\
                            &PDF   &0.87s  &0.85s  &0.88s  &0.88s   \\
            \midrule
                            &SOM  &91.10s  &90.15s  &91.59s  &89.70s   \\ 
                            &FBE-CIE  &257.02s  &284.69s  &253.37s  &259.06s   \\
            $\epsilon_r=8$  &uSOM  &79s  &80s  &78s  &79s   \\
                            &PDNN  &321s  &235s  &240s  &253s   \\
                            &PDF   &0.87s  &0.88s  &0.87s  &0.88s   \\
            \bottomrule
        \end{tabular}
        }
    \end{table}
    
    \subsection{Effectiveness of CCO and $L^\mathrm{Bridge}$}
    \label{subsec:ContrCCOandBri}
    \begin{figure}[!t]
    	\centering
    	\includegraphics[width = .5\linewidth]{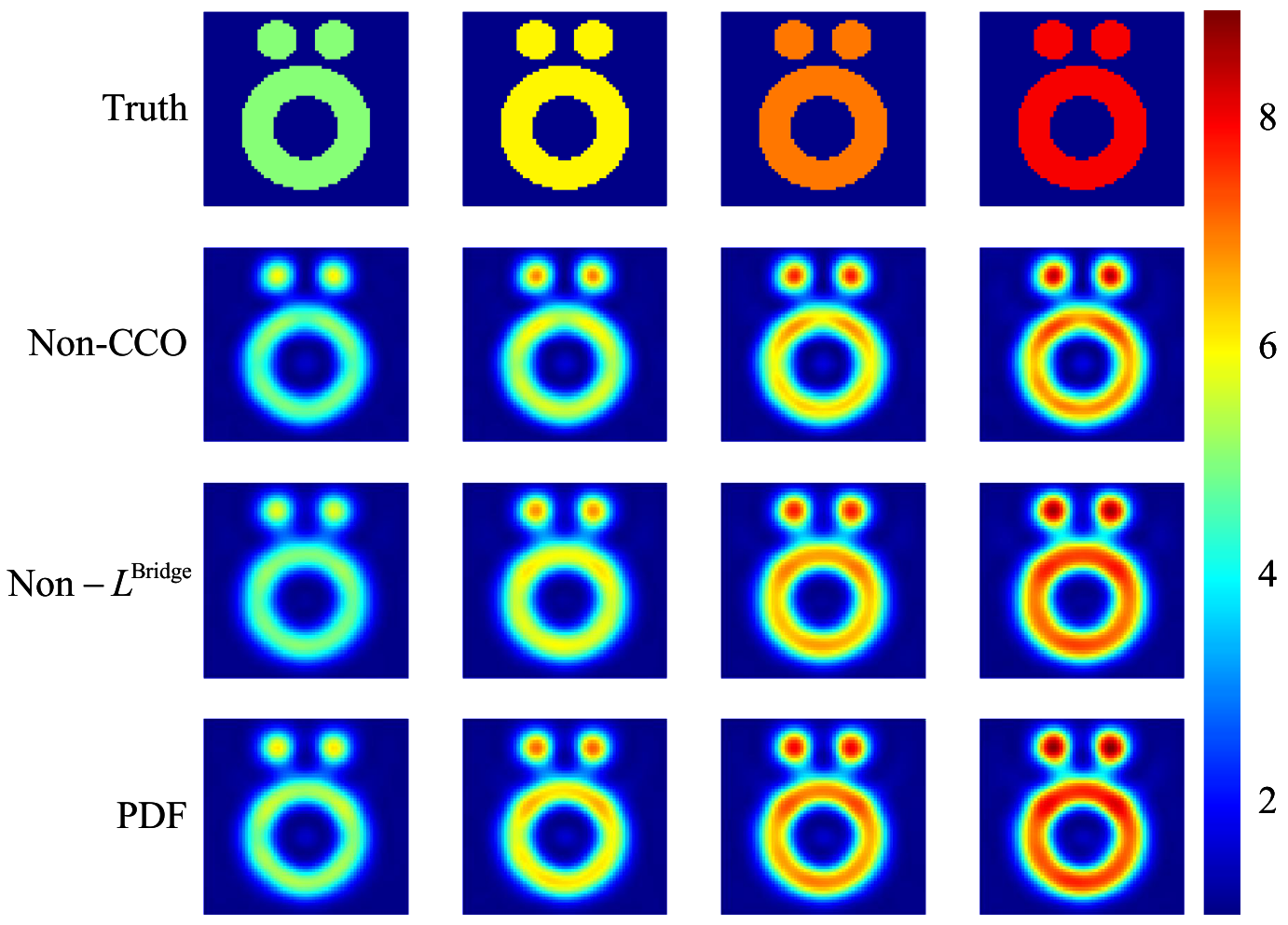}
    	\caption{The reconstruction results of PDF and without the proposed contrast-compensated Operator (CCO) or $L^\mathrm{Bridge}$ for different relative permittivity Austria profiles.}
    	\label{fig:ContrCCOandBri}
    \end{figure}
    The quantitative accuracy and spatial resolution of the proposed PDF solver are significantly enhanced by the integration of the contrast-compensated operator (CCO) and the specialized bridge-suppressing loss term. As discussed in Section~\ref{subsec:CCOperator}, reconstructions based on a truncated Fourier basis often exhibit a characteristic ``edge roll-off" effect, where the permittivity values are correctly estimated in the object’s interior but gradually diminish toward its boundaries. This systematic attenuation is a direct consequence of omitting high-frequency spectral components, which are necessary to represent sharp dielectric interfaces. By applying the CCO as a post-processing step, we can adaptively compensate for this spectral-domain-induced attenuation. As illustrated in Figure~\ref{fig:ContrCCOandBri}, the inclusion of the CCO restores the peak permittivity values, bringing the reconstruction significantly closer to the ground truth while sharpening the object boundaries across various contrast levels.
    
    Simultaneously, the solver addresses a common resolution bottleneck in inverse scattering: the tendency of iterative algorithms to produce artificial ``bridges" or non-physical connections between closely spaced scatterers. This phenomenon is particularly problematic when the mutual coupling between neighboring targets is strong, leading to a loss of target separability. To alleviate this, the proposed $L^{\text{Bridge}}$ term is introduced to penalize regions of high amplitude with low local gradients, effectively acting as a spatial gate that discourages the formation of these artifacts. The comparative results in Figure~\ref{fig:ContrCCOandBri} demonstrate that with the incorporation of $L^{\text{Bridge}}$, the artificial connections between the circular and annular components of the Austria profile are effectively alleviated. 
    
    
    \subsection{Ablation Study on $L^\mathrm{Bound}$ and $L^\mathrm{TV}$}
    \label{subsec:ContrLBouandLTV}
    
    To assess the individual contributions of the physical boundary constraint ($L^\text{Bound}$) and the total variation regularization ($L^\text{TV}$), an ablation study is performed on a high-contrast Austria profile ($\epsilon_r = 8$). The solver's performance is evaluated under a range of conditions, from noise-free measurements to a severe SNR$ = 1$ dB environment. By selectively removing these terms while maintaining all other network parameters and training configurations, we can isolate their impact on the stability and physical reliability of the reconstructed permittivity distributions.
    
    \begin{figure}[!t]
    	\centering
    	\includegraphics[width = .6\linewidth]{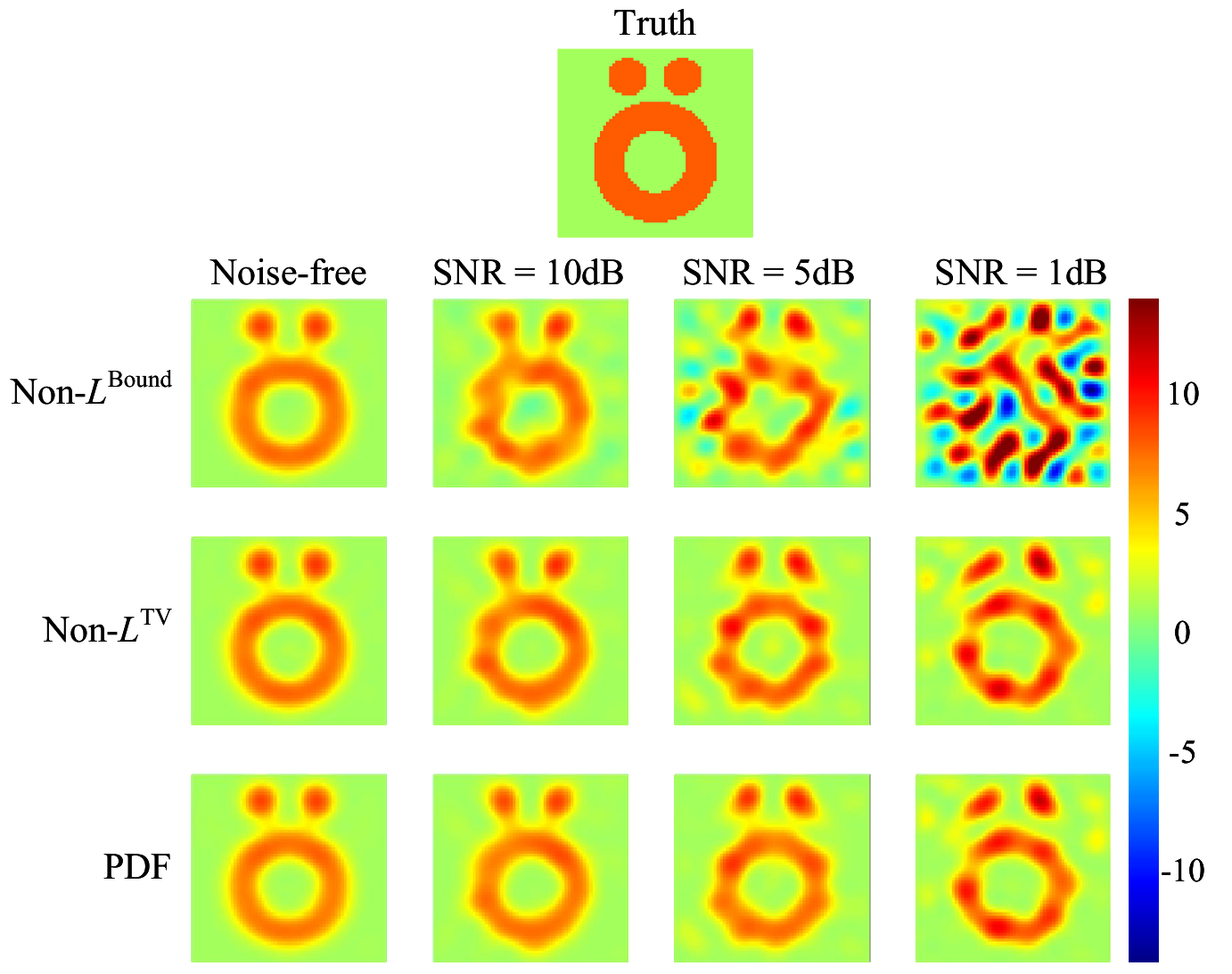}
    	\caption{The imaging results of PDF and without $L^\mathrm{Bound}$ or $L^\mathrm{TV}$ are compared under noise-free conditions and additive noise levels of SNR = 10dB, 5dB, and 1dB.}
    	\label{fig:ContrLBouandTV}
    \end{figure}
    The boundary constraint $L^\text{Bound}$ is designed to enforce the fundamental physical requirement that the real part of the relative permittivity in a dielectric medium must be greater than or equal to that of free space (i.e., $\text{Re}\{\epsilon_r\} \ge 1$). As illustrated in Figure~\ref{fig:ContrLBouandTV}, in the absence of noise, the removal of $L^\text{Bound}$ has negligible impact on the reconstruction, as the physics-driven optimization naturally tends toward physically plausible solutions when provided with clean data. However, as the noise level increases, the inversion process becomes increasingly ill-posed, and the optimizer may minimize the data misfit by introducing non-physical, negative permittivity values in the background. Without $L^\text{Bound}$, the reconstruction completely fails at SNR$ = 1$ dB, yielding a solution dominated by non-physical oscillations. In contrast, the PDF solver with the boundary constraint effectively suppresses these artifacts, maintaining a clean background and ensuring that the reconstructed profile remains within the physical bounds of common dielectric media.
    
    Complementing the boundary constraint, the $L^\text{TV}$ term acts as a spatial regularizer that promotes piecewise smoothness while preserving the sharp dielectric interfaces of the scatterers. The results shown in Figure~\ref{fig:ContrLBouandTV} indicate that $L^\text{TV}$ is particularly effective at mitigating excessive local overestimation and high-frequency noise fluctuations within the reconstructed objects. By penalizing the $L_1$-norm of the gradient, $L^\text{TV}$ suppresses the granular artifacts that often plague unregularized neural network solvers, leading to a more homogeneous and structurally consistent permittivity distribution. 
    
    The synergy between these two terms is therefore essential. $L^\text{Bound}$ ensures global physical consistency and robustness against severe noise, while $L^\text{TV}$ provides the necessary spatial regularization to achieve a stable and clean reconstruction.
    
    \subsection{Performance on Complex and Benchmarked Datasets}
    \label{subsec:reconComSca}
    
    \begin{figure}[!t]
    	\centering
    	\includegraphics[width = .56\linewidth]{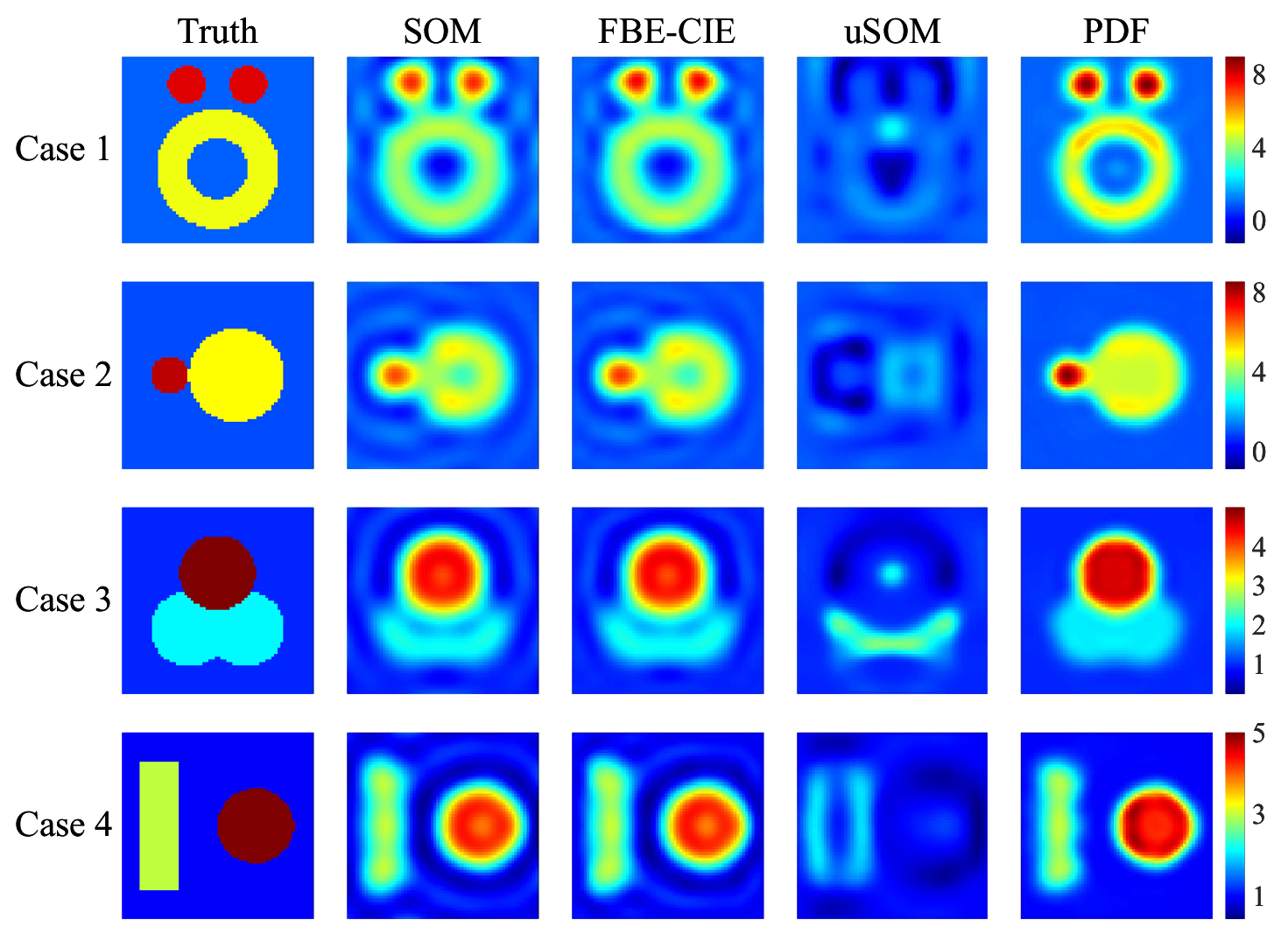}
    	\caption{Comparison of imaging results for complex scatterers obtained by SOM, FBE-CIE, uSOM and PDF solver.}
    	\label{fig:CompSca}
    \end{figure}
    
    To further demonstrate the versatility and generalization capability of the proposed PDF solver, we extend our evaluation beyond the standard Austria profile to a series of more challenging geometries, denoted as Case 1 through 4. These cases include sharp-edged polygons, high-contrast overlapping cylinders, and targets with intricate concave features that intensify the effects of mutual coupling and multiple scattering. As illustrated in Figure~\ref{fig:CompSca}, the PDF solver successfully resolves these complex configurations with high spatial fidelity. Notably, while the Fourier-spectral basis is inherently smooth, the synergy between the bridge-suppressing loss and the TV regularization allows the solver to capture the sharp dielectric interfaces of polygonal objects without the severe Gibbs ringing artifacts that typically plague spectral methods. This performance indicates that the solver is not only robust but also capable of preserving the structural integrity of diverse targets, a requirement that is crucial for practical industrial and security screening applications.

    \subsection{Uncertainty Analysis}
    \label{subsec:UncAna}
    
    \begin{figure}[!t]
    	\centering
    	\includegraphics[width = .6\linewidth]{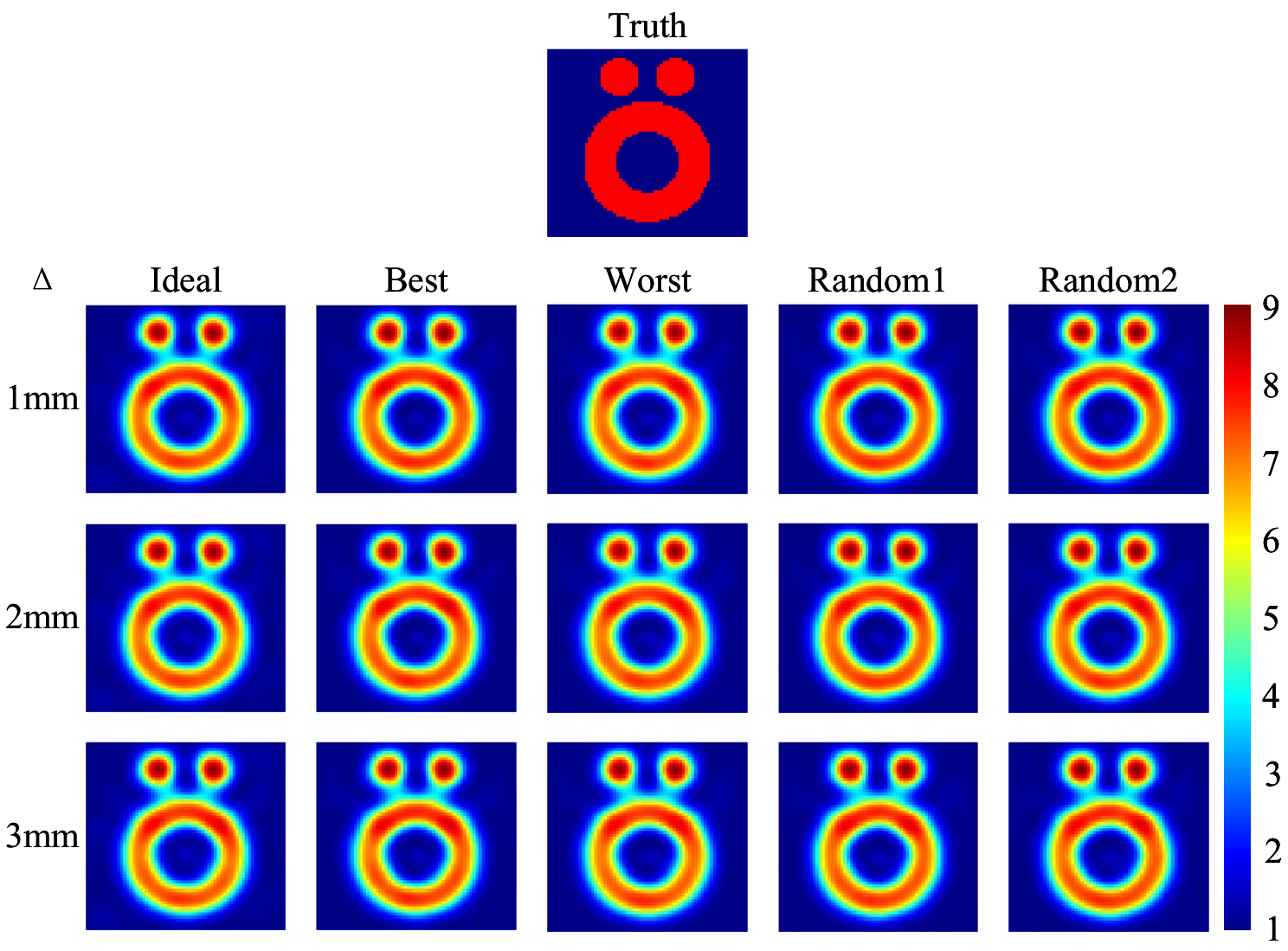}
    	\caption{The imaging results under antenna position uncertainties with no position perturbation, the best, worst, and two randomly selected reconstruction among the 100 realizations with antenna position perturbations.}
    	\label{fig:UnaAna}
    \end{figure}
    
    \begin{figure}[!t]
    	\centering
    	\includegraphics[width = 0.5\linewidth]{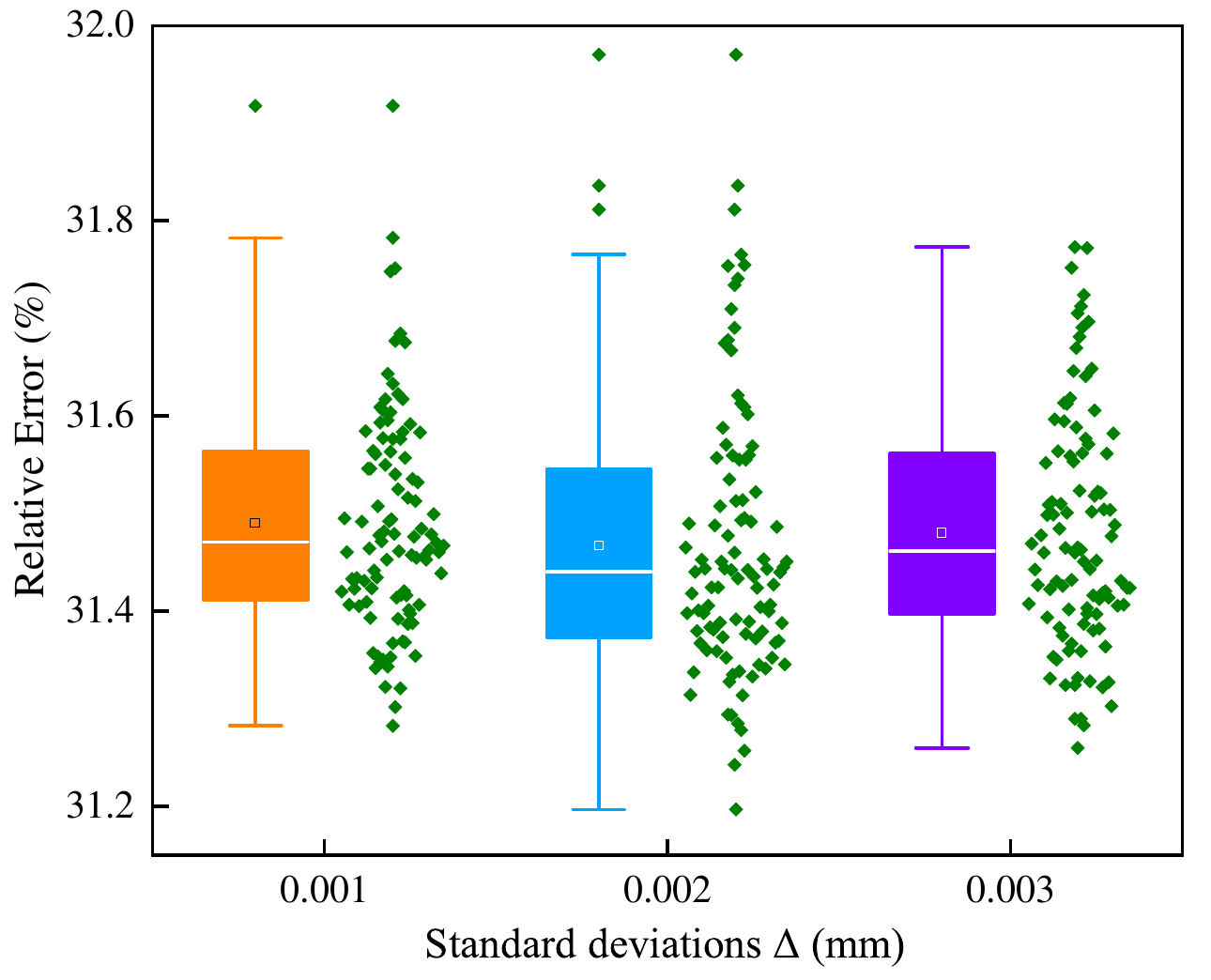}
    	\caption{Boxplots of the relative errors under antenna position perturbations with Gaussian standard deviations  $\Delta=1\mathrm{mm}$, $2\mathrm{mm}$ and $3\mathrm{mm}$.}
    	\label{fig:boxplot}
    \end{figure}
    
    In practical microwave imaging systems, uncertainties in antenna positioning are inevitable due to mechanical tolerances, thermal expansion, or manual alignment errors. To evaluate the robustness of the proposed PDF solver against these real-world perturbations, we perform a statistical uncertainty analysis by introducing perturbations to the coordinates of both transmitters and receivers. Perturbations following the Gaussian distribution with a zero mean and standard deviations of $\Delta = 1$ mm, $2$ mm, and $3$ mm are applied to the antenna locations. For each setting of deviation, 100 independent realizations are simulated to characterize the resulting error distribution, providing an assessment of the solver’s stability under non-ideal measurement conditions.
    
    Visual evidence of this robustness is provided in Figure~\ref{fig:UnaAna}, which compares the reconstruction results for the best-case, worst-case, and randomly selected realizations among the 100 samples. Even in the worst-case scenarios under maximum perturbation, the PDF solver successfully preserves the overall morphology and contrast of the targets, with only minor background fluctuations appearing. This performance highlights the practical utility of the proposed method. 
    
    The statistical results, summarized in the boxplot of Figure~\ref{fig:boxplot}, reveal that the PDF solver exhibits remarkable resilience to position inaccuracies. At a perturbation level of $\Delta = 1$ mm, the variance between the maximum and minimum relative reconstruction errors is restricted to approximately $0.6\%$. Even as the standard deviation increases to $3$ mm—a significant displacement relative to the operating frequency—the error distribution remains tightly clustered, with no evidence of catastrophic divergence or extreme outliers. This high degree of stability suggests that the Fourier-spectral parameterization and the CIE-based formulation effectively act as an implicit filter, preventing small-scale geometric errors in the measurement setup from being amplified into large-scale artifacts in the reconstructed permittivity profile. Consequently, the proposed framework is well-suited for high-fidelity imaging in field-deployable systems where precise antenna calibration and maintenance are difficult to guarantee.
    
    \section{Experimental Validation}
    \label{sec:expVal}
    
    \begin{figure}[!t]
    	\centering
    	\includegraphics[width = .6\linewidth]{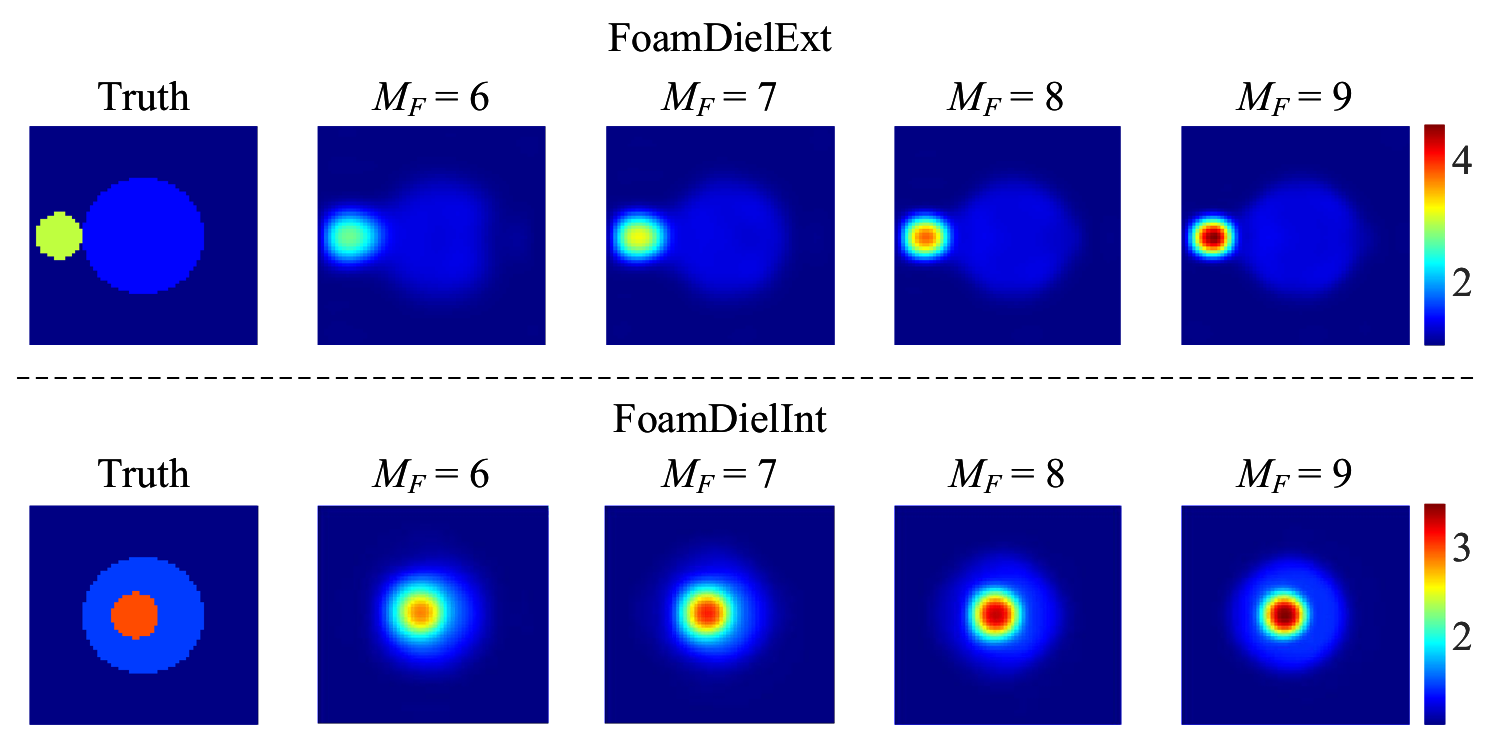}
    	\caption{The imaging results of the proposed PDF solver using the experimental data "FoamDielExt" and "FoamDielInt" under different numbers of low-frequency Fourier bases $M_F = 6, 7, 8,$ and $9$, with $\beta$ fixed to 6.}
    	\label{fig:FresnelExperiment}
    \end{figure}
    
    The practical efficacy of the proposed PDF solver is evaluated using experimental datasets provided by the Fresnel Institute \cite{geffrin2005Fresnel}, specifically the ``FoamDielExt" and ``FoamDielInt" cases. Transitioning from synthetic data to experimental measurements introduces significant challenges, including antenna mutual coupling, cable-induced phase instabilities, and positioning inaccuracies that are not perfectly captured by idealized numerical models. Despite these non-ideal factors, the PDF solver demonstrates remarkable robustness. By embedding the contraction integral equation (CIE) directly into the optimization loop, the solver effectively mitigates the nonlinear scattering effects inherent in these physical measurement setups. As shown in Figure~\ref{fig:FresnelExperiment}, the reconstructed permittivity profiles for the foam and dielectric cylinders are quantitatively accurate and exhibit a clean background, maintaining high structural fidelity even in the presence of real-world measurement uncertainties.
    
    A key focus of this experimental analysis is the sensitivity of the solver to the Fourier truncation order ($M_F$) when dealing with actual measured fields. As illustrated in the reconstruction results, the PDF solver maintains a stable performance across various $M_F$ configurations ($M_F = 6, 7, 8, 9$). While a lower $M_F$ value like 6 results in a slightly smoother profile due to the implicit low-pass filtering, it remains highly effective at suppressing experimental noise. Conversely, higher $M_F$ values provide sharper boundaries but are more susceptible to the amplification of experimental perturbations. The results confirm that $M_F = 7$ provides a reliable balance for experimental data, successfully recovering both the moderate-contrast foam ($\epsilon_r \approx 1.45$) and the high-contrast dielectric cylinders ($\epsilon_r \approx 3.0$) without introducing significant artifacts.
    
    
    \section{Conclusions}
    \label{sec:conclusions}
    In this paper, we have proposed a real-time physics-driven Fourier-spectral (PDF) neural network solver designed to overcome the computational bottlenecks of traditional untrained neural network solvers for inverse scattering problems. By leveraging the spectral compressibility of the induced contrast source and representing it within a truncated Fourier basis, the PDF solver effectively reduces the dimensionality of the optimization space. This spectral-domain approach enables the solver to achieve sub-second reconstruction, a nearly 100-fold speedup compared to existing untrained solvers.
    
    To ensure high-fidelity reconstructions in the presence of strong multiple scattering, the solver incorporates the contraction integral equation to mitigate nonlinearity and a specialized bridge-suppressing loss to enhance spatial resolution between closely spaced targets. Furthermore, the introduction of a contrast-compensated operator effectively corrects for the systematic contrast attenuation inherent in truncated spectral representations, ensuring quantitative accuracy across a wide range of dielectric profiles.
    
    Numerical simulations and experimental validations demonstrate that the PDF solver is not only highly efficient but also remarkably robust to severe noise and antenna positioning uncertainties. The solver consistently outperforms state-of-the-art benchmarks in terms of artifact suppression and structural fidelity, particularly in high-contrast and complex geometric scenarios. By bridging the gap between the speed of data-driven deep learning and the physical consistency of iterative solvers, the proposed PDF framework offers a promising solution for real-time microwave tomography in medical imaging, non-destructive testing, and security screening. Future work will focus on extending this spectral-domain physics-driven approach to three-dimensional imaging and exploring its potential in adaptive measurement configurations.
	
	\bibliographystyle{unsrt}
	\bibliography{References}
\end{document}